\documentclass[10pt,twocolumn,letterpaper]{article}

\usepackage[pagenumbers]{wacv} 

\usepackage{graphicx}
\usepackage{amsmath}
\usepackage{amssymb}
\usepackage{booktabs}
\usepackage{siunitx}
\usepackage{tabularray}
\usepackage{float}
\usepackage[super]{nth}

\usepackage[pagebackref,breaklinks,colorlinks]{hyperref}

\usepackage[capitalize]{cleveref}
\crefname{section}{Sec.}{Secs.}
\Crefname{section}{Section}{Sections}
\Crefname{table}{Table}{Tables}
\crefname{table}{Tab.}{Tabs.}

\def\wacvPaperID{2} 
\def\confName{WACV ULTRRA Workshop}
\def\confYear{2025}

\begin{document}

\title{Metadata-free Georegistration of Ground and Airborne Imagery}

\author{Adam Bredvik \\
\and Scott Richardson \\
\\
Vision Systems, Inc.\\
{\tt\small visionsystemsinc.com} \\
\and Daniel Crispell 
}
\twocolumn[{%
\renewcommand\twocolumn[1][]{#1}%
\maketitle
\centering
\captionsetup{type=figure}
\vspace{-15mm}
\includegraphics[width=\linewidth]{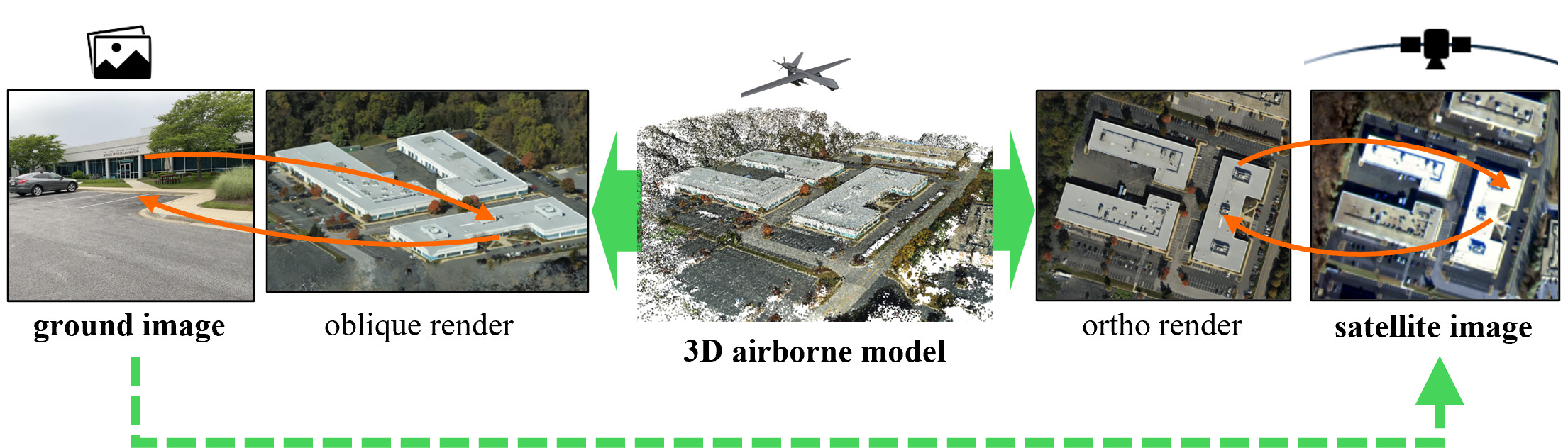}
\captionof{figure}{The proposed georegistration method uses synthetic NeRF renderings generated from airborne imagery to bridge the gap between ground-level and satellite imagery. NeRF rendering allows the control necessary to produce images suitable for 2D matching to both satellite and ground-level imagery, allowing the transfer of satellite geopositioning to ground level images without metadata. Satellite imagery (C) 2025 Maxar}
\vspace*{0.45cm}
}]


{ 
\raggedbottom

\begin{abstract}
Heterogeneous collections of ground and airborne imagery can readily be used to create high-quality 3D models and novel viewpoint renderings of the observed scene.
Standard photogrammetry pipelines generate models in arbitrary coordinate systems, which is problematic for applications that require georegistered models.
Even for applications that do not require georegistered models, georegistration is useful as a mechanism for aligning multiple disconnected models generated from non-overlapping data.
The proposed method leverages satellite imagery, an associated digital surface model (DSM), and the novel view generation capabilities of modern 3D modeling techniques (\eg neural radiance fields) to provide a robust method for georegistering airborne imagery, and a related technique for registering ground-based imagery to models created from airborne imagery.
Experiments demonstrate successful georegistration of airborne and ground-based photogrammetric models across a variety of distinct sites.
The proposed method does not require use of any metadata other than a satellite-based reference product and therefore has general applicability.
\end{abstract}


\section{Introduction}
\label{sec:intro}
Standard photogrammetic toolchains can create 3D models from collections of 2D images, but without prior knowledge of the sensor positions the generated 3D models are ambiguous up to a scale, translation, and rotation in the best case~\cite{hartley_zisserman}.
The primary goal of georegistration is to estimate and apply a transform to a 3D model that brings it into a coordinate system with known absolute position, orientation, and scale.
The absolute positioning of each individual model also provides relative positioning of individual models that are independently georegistered, which is critical when models are built from non-overlapping sets of input imagery.
Finally, georegistration of 3D models also provides precise positioning of the sensors used to capture the imagery used to create the photogrammetric models.

The proposed method leverages the novel-view generation capability of neural radiance fields (NeRFs) to create synthetic viewpoints of the modeled region that are more readily matched across domains (\eg airborne to satellite/ground-level imagery) and include per-pixel estimates of 3D location, which is used to convert the 2D matches to a 3D alignment.
The proposed method assumes the availability of a satellite orthophoto and digital surface model (DSM) overlapping the reconstructed area of interest.

The remainder of the paper is laid out as follows.
Related methods are discussed in \cref{sec:related_work}.
The proposed method is described in detail in \cref{sec:method}, and experimental results are presented in \cref{sec:results}.
Final conclusions and identification of future improvements to the methods are discussed in \cref{sec:conclusion}.

} 

\section{Related work}
\label{sec:related_work}
\subsection{Airborne to satellite registration}

Georegistration of airborne imagery by identifying correspondences in the collection of images with a satellite-centric view has been studied in various forms for over 20 years \cite{sheikh2004, wu2006, Jaehong2010}.
In 2010, for example, Jaehong, \etal~\cite{Jaehong2010} used SIFT keypoint matching~\cite{lowe_sift_2004} to georegister airborne imagery to satellite stereo pairs.


Because keypoint matching is the cornerstone of georegistration, SIFT's limited robustness to changes in appearance and viewpoint between images restricts the range of airborne images that can be georegistered.
In 2023, Liu \etal~\cite{liu2023} adapted the traditional pipeline to georegister airborne images from unmanned aerial vehicles (UAVs) using SuperPoint/SuperGlue \cite{sarlin2020superglue}, a trained feature detector and matcher.
Berton \etal~\cite{berton2024} used the detector-free dense matcher, RoMa \cite{roma}, 
and Bokman \etal~used a rotation equivariant version of DeDoDe \cite{edstedt2024} keypoint matching
to find correspondences between satellite images with impressive results.

Similar to this work, Krock \etal~\cite{krock2022} map an airborne structure from motion (SfM) reconstruction from relative to world coordinates. Their pipeline depends on matches between a satellite image and the orthorectified ground plane of an airborne image -- which can induce significant distortion, making matching harder.
To facilitate matching, we instead \emph{render} a synthetic (nadir) view of the airborne model. While this rendered image has less distortion, a synthetic image can, nonetheless, be difficult to match due to scale, appearance differences, and artifacts due to the modeling and rendering pipeline.
For this reason, we follow Berton \etal~\cite{berton2024} and use RoMa to identify matches between the satellite and rendered airborne image.

Finally, in contrast to the approaches of Krock \etal~\cite{krock2022}
and Zhuo \etal~\cite{zhuo2017}, our proposed method is metadata-free and does not rely on measurements from onboard sensors such as from inertial measurement units (IMU) or GPS for georegistration.
In fact, Krock \etal~assume they already have precomputed cameras and 3D model (in a local coordinate system). Instead, we generate a 3D model and associated cameras given only the airborne imagery. 



\subsection{Ground to airborne registration}


Several existing approaches~\cite{pro2024semantic, lin2015learning, workman2015localize, zhu2022transgeo, Deuser_2023_ICCV, Fervers_2023_CVPR} attempt to directly match overhead and ground-level views via descriptors learned from many paired overhead/ground training examples.
We instead focus on leveraging 3D modeling and rendering to bridge the domain gap, which does not require any domain-specific training data.

MeshLoc~\cite{Panek2022ECCV} replaces the typical sparse feature-based approach to visual localization with image-based matching between real and rendered imagery. Unlike the proposed appraoch, Meshloc is focused on matching within a single domain.

Other previous approaches to geometry-aided cross-domain matching rely on approximate geolocation metadata to form initial matches.
Shan \etal~\cite{shan2014accurate} warp ground views using depth maps to facilitate direct matching to airborne imagery using approximate position and orientation information from both images.
Zhu \etal~\cite{zhu2020leveraging} leverage rendered 3D models to aid airborne to ground matching, but assume an initial rough registration using metadata.
Yan \etal~\cite{yan2023render} propose an iterative ``render and compare'' approach to cross-domain image matching, initialized with coarse geolocation from GPS. 
The proposed approach assumes that the input imagery is localized to a coarse area of interest, but makes no assumption about metadata in the ground or airborne imagery.

\section{Metadata-free georegistration}
The following section covers our metadata-free method for georegistering airborne and ground-based imagery which, by rendering novel views of the airborne model (\ref{sec:novelview}), establishes robust correspondences (\ref{sec:robustcorr}) that bring (first) the airborne (\ref{sec:airborne_georeg}) and then the ground-level imagery (\ref{sec:ground_georeg}) into alignment with the satellite imagery.

\label{sec:method}

\subsection{Novel view rendering}
\label{sec:novelview}
The proposed georegistration pipeline leverages the ability of NeRF~\cite{mildenhall_nerf_2020} reconstruction methods to generate photorealistic novel view renderings to ``bridge the gap'' between dominant viewpoints in different domains (ground, airborne, and satellite), making the 2D matching problem easier and more robust.
The following subsection describes the pipeline used to reconstruct 3D models and render novel views given a set of unposed input images.
The ground and airborne image sets are reconstructed independently and merged only through the georegistration process described in \cref{sec:robustcorr} and \cref{sec:georegistration}.

\subsubsection*{Structure from motion}
\label{sec:sfm}

Input images are assumed to be metadata-free (\ie unposed), and therefore require calibration and pose estimation.
A structure from motion (SfM) pipeline composed of feature detection, feature matching, and bundle adjustment generates estimates of camera intrinsic parameters (focal length, lens distortion), extrinsic parameters (position and orientation), and 3D locations of all reconstructed feature points.
DeDoDe~\cite{dedode} and RoMa~\cite{roma} detect and match keypoints across the input images, followed by ``doppelganger'' detection~\cite{doppelgangers} to eliminate challenging visually similar but geometrically distinct false image matches.
COLMAP's~\cite{schoenberger2016sfm} incremental bundle adjustment then generates estimates for camera metadata and 3D feature point locations.
Multiple independent reconstructions are reported in cases where COLMAP is unable to create a single sparse reconstruction for all the input images due to a lack of good matches between the images.
A reconstruction is sometimes referred to as a subgraph defined by the cameras connected by co-visible 3D points.
Each reconstruction provided by COLMAP exists in an arbitrary coordinate system without well-defined scale, orientation, or location.

\subsubsection*{Gravity vector estimation}
Structure-from-motion (SfM) pipelines such as COLMAP produce sparse reconstructions in an arbitrary coordinate system that must be oriented to a known reference frame before one can render nadir and oblique views of the scene.
Without metadata containing camera orientation, a combination of the original imagery and the reconstructed 3D geometry is used to determine the gravity vector.
The ground is assumed locally flat, and a plane is fit to the set of reconstructed points corresponding to ground-labeled pixels in the input images.
The gravity vector is identified as the upward-facing surface normal to the estimated ground plane.
The pretrained OneFormer model~\cite{jain2023oneformer} for universal image segmentation is used to identify which points correspond to ground pixels.
All ground-like categories (\eg road, grass, water) are grouped together to form a single binary mask that is used to identify 3D ground points.

A simple RANSAC-based \cite{RANSAC} algorithm is used for fitting a plane to these points, which results in two possible normal vectors.
To determine which vector points ``up'', we enforce the constraint that rays originating from the cameras must intersect the ``top'' side of the plane.
The majority of these rays should produce a negative dot product with the true gravity vector, since the angle created between them at the point of intersection should be larger than 90°.
Once the gravity vector is known, the site model is then rotated such that the local $z$ axis points up.

\begin{figure}
    \centering
    \includegraphics[width=\linewidth]{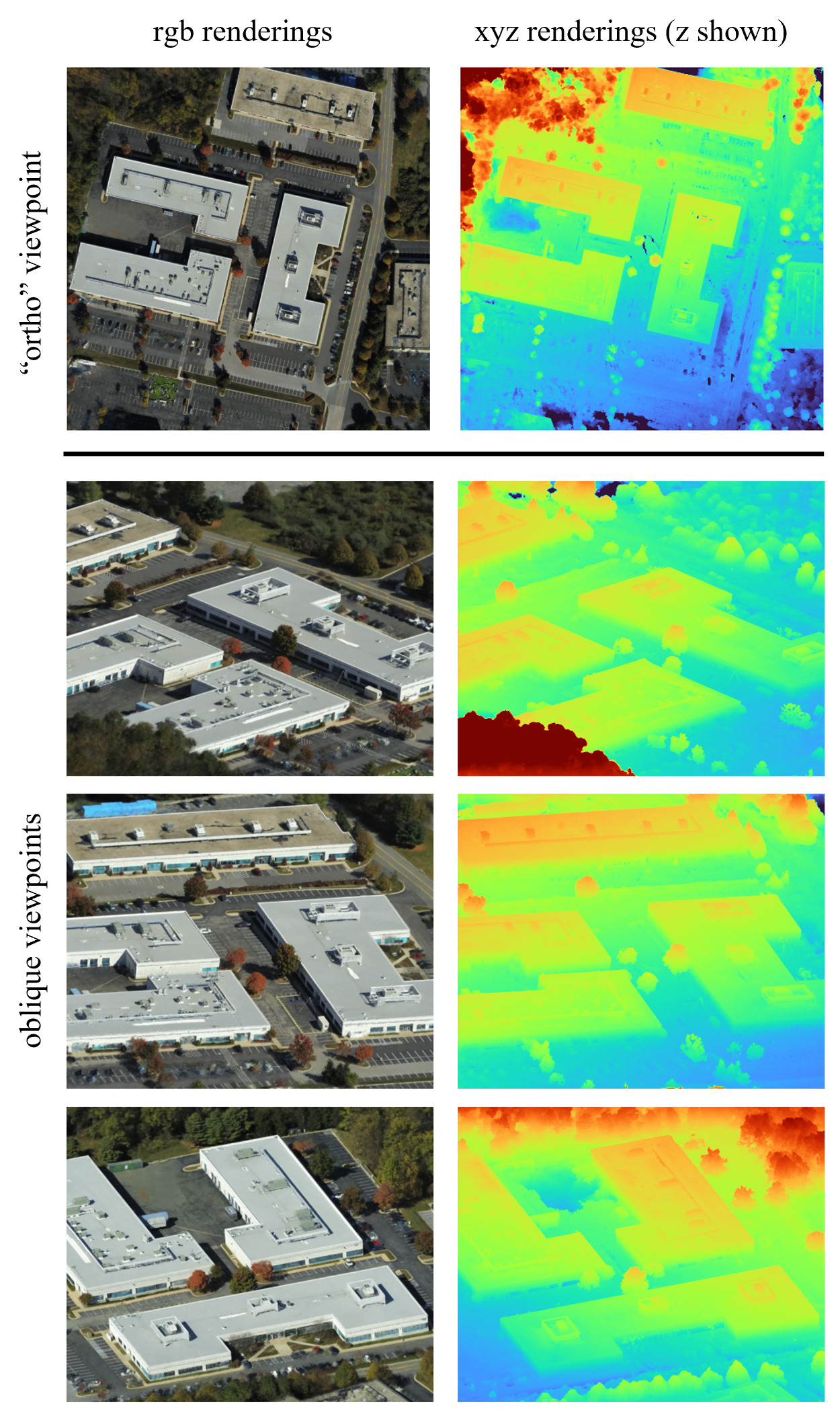}
    \caption{The 3D airborne model is used to render an nadir-view orthographic image (top) and a series of oblique renderings at regular azimuth spacings. Per-pixel 3D point estimates (right) and uncertainty (not shown) are rendered for each image.}
    \label{fig:renders_grid}
\end{figure}

\subsubsection*{NeRF reconstruction and rendering}
Neural Radiance Fields~\cite{mildenhall2021nerf} (NeRF) represent 3D scenes using a volumetric density function that is integrated along viewing camera rays to produce rendered color.
The proposed approach uses a variation of the ``nerfacto-huge'' model as implemented in Nerfstudio~\cite{nerfstudio} to reconstruct and render a 3D model from the input airborne imagery and associated COLMAP-estimated cameras (\cref{sec:sfm} § Structure from motion).
In addition to the standard visual reconstruction loss, the optimization is modulated by per-pixel depth maps generated by COLMAP's dense stereo matching algorithm~\cite{schoenberger2016mvs} (absolute depth constraint) and single-image monocular depth estimation~\cite{marigold} (relative depth constraint~\cite{guangcong_sparsenerf_2023}).
The depth-based geometric regularization improves reconstruction and novel view rendering quality, especially on textureless surfaces such as pavement and rooftops.

Once reconstructed, the NeRF model is used to render a series of synthetic images that are used downstream for matching to the satellite and ground-level images.
The 2D image matches must be lifted to 3D in order to compute the georegistration transformations, so per-pixel depth maps are rendered in addition to the color images.
The ``Bayes' Rays''~\cite{goli_bayes_2024} uncertainty quantification method produces a per-pixel confidence value used to unreliable matches.
The synthetic viewpoints generated for matching to the airborne and ground-level images are shown in \cref{fig:renders_grid} and described in \cref{sec:airborne_georeg} and \cref{sec:ground_georeg}, respectively.

\subsection{Robust correspondence identification}
\label{sec:robustcorr}
The key to establishing registration between the target NeRF models and source imagery is the generation of reliable image matches between the rendered and source images.
Modern dense-matching techniques~\cite{roma, dkm, pdc-net+} are trained to produce matches between images that come from radically different views, captured at different times, and under different environmental conditions.
This leads to large appearance transformations between the images and forces the network to be robust to such changes.

We use the RoMa~\cite{roma} matcher, which leverages the frozen pretrained (coarse) features from the DINOv2~\cite{oquab_dinov2_2024} foundation model combined with specialized ConvNet-based fine features, creating a localizable feature pyramid.
Matches are made using a Gaussian Process match encoder and a transformer match decoder (without position encodings) in a coarse-to-fine manner.

RoMa can identify matches between images that vary in appearance due to shadows, season, construction, transient objects, sensor, viewpoint, scale, and even between real and synthetic (rendered) images.

\subsection*{Cyclic Consistency}
\label{sec:denseflow}
RoMa is a detector-free dense matcher.
In contrast to sparse matchers like SIFT,
dense matchers produce a dense flow in which every point in one frame is matched with its corresponding point in the next.

\begin{figure*}
    \centering
    \includegraphics[width=0.9\linewidth]{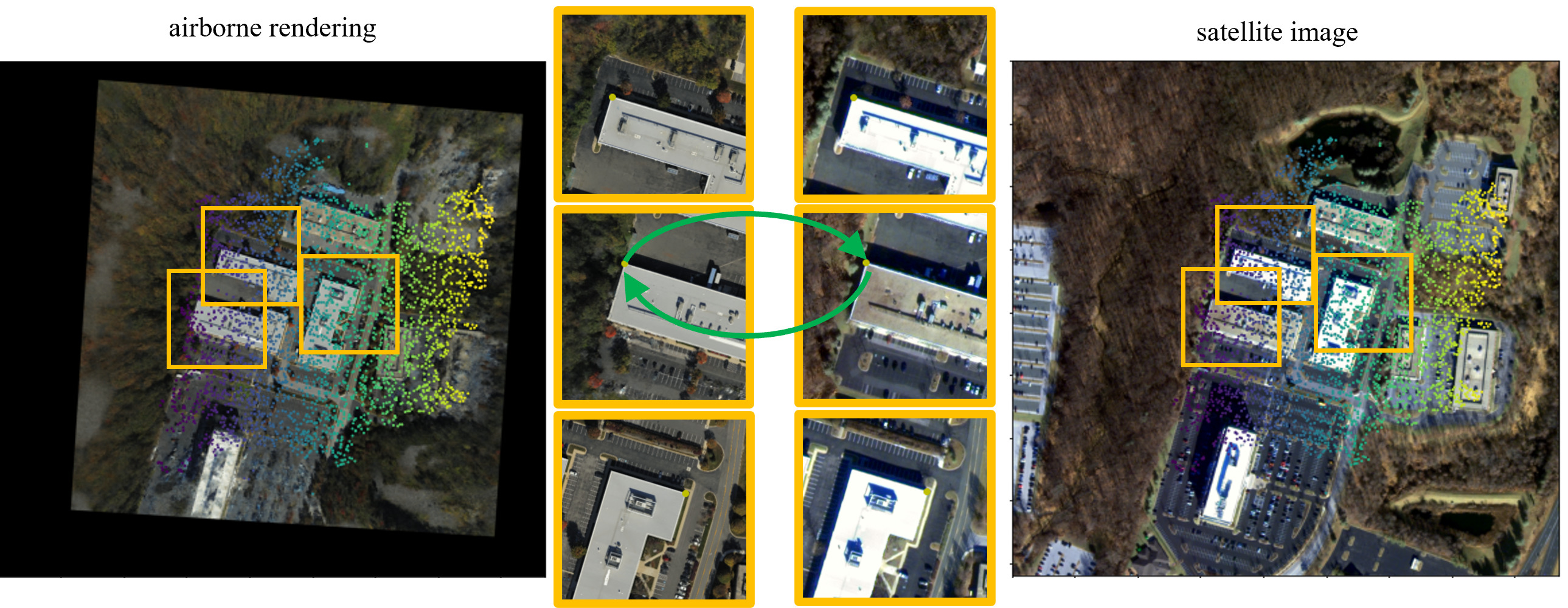}
    \caption{Dense matching with cyclic consistency results in high-quality reliable matches between the rendered airborne models and satellite imagery. Satellite imagery (C) 2025 Maxar}
    \label{fig:airborne_matching}
\end{figure*}

RoMa computes (independently) both the dense flow (with shape $[H',W']$) that maps image A (with shape $[H,W]$) to image B (with shape $[H',W']$), as well as the dense flow (with shape $[H,W]$) that maps image B to image A; these two flows are sometimes referred to as the forward and backward dense flow (or warp).
A match defined by these two flows is \emph{cyclically consistent} when a point $\mathbf{x}$ in image A that matches $\mathbf{y}$ in image B, according to the forward flow, and point $\mathbf{y}$ in image B matches $\mathbf{x}$ in image A according to the reverse flow.
\begin{equation}
    \label{eq:cyclic}
    \begin{split}
    F(F^{-1}(\mathbf{x})) & = F(R(\mathbf{x})) \approx \mathbf{y} \\
    R(R^{-1}(\mathbf{y})) & = R(F(\mathbf{y})) \approx \mathbf{x} \\
    \end{split}
\end{equation}
Cyclic consistency is measured as $c = \| F(R(\mathbf{x})) - \mathbf{x} \|$ and any matches with $c > N$ are discarded.
We use an $N=2$ pixels.
In addition to cyclic consistency, the RoMa matcher itself generates a per-pixel confidence map.
We discard any matches with confidence $< 0.2$.
If a match is cyclically consistent, it is likely (but not always) a correct match.
Examples of cyclically consistent matches for airborne and ground images are shown in \cref{fig:airborne_matching} and \cref{fig:ground_tile_matching}, respectively.

\subsection{Georegistration}
\label{sec:georegistration}

\subsubsection*{Lifting 2D matches to 3D points}
Given the 2D matches between either the ground and airborne rendered image or the airborne rendered image and the satellite DSM, the 2D matches are lifted to corresponding 3D points.
In both cases, an accompanying map defines the 3D location (in some coordinate frame) of each pixel in the respective image.

In the case of the airborne-to-satellite DSM georegistration, a co-rendered ``xyz image'' (per-pixel 3D coordinates) defines the 3D coordinates (wrt a local coordinate frame) of the rendered aerial image, and a co-registered satellite DSM defines the 3D coordinates (wrt the WGS84 datum) of the satellite orthophoto.

Similarly, in the case of the ground-to-airborne georegistration, the rendered airborne image has a co-rendered xyz image, and the ground image has an xyz image produced from a COLMAP dense SfM pre-computed from the site's ground-level imagery (\cref{sec:sfm} § Structure from motion).

\subsubsection*{Compute 3D Transforms}
\label{sec:compute_transform}

Given a set of corresponding 3D points, a similarity transform is estimated using RANSAC and the Kabsch-Umeyama method~\cite{lawrence2019purely}, which computes a least-squares estimation of the transformation’s parameters.
First, the 3D-to-3D correspondences between the airborne render and the satellite reference are used to compute a 3D similarity transform that maps the airborne 3D model into the DSM’s coordinate reference frame (\eg WGS84), thereby georegistering the model.
Next, a second 3D similarity transform is computed that transforms the ground 3D model(s) to the coordinate frame of the georegistered airborne model.
The coregistered ground, airborne, and satellite models are shown in \cref{fig:three_sites}.

\subsection{Georegistration of airborne imagery}
\label{sec:airborne_georeg}

Metadata-free georegistration of airborne imagery aims to transform the airborne model -- which, due to the metadata-free nature of the source imagery, is constructed in an arbitrary coordinate frame -- into the georeferenced coordinate frame of the satellite-based reference data.

\begin{figure}
    \centering
    \includegraphics[width=0.8\linewidth]{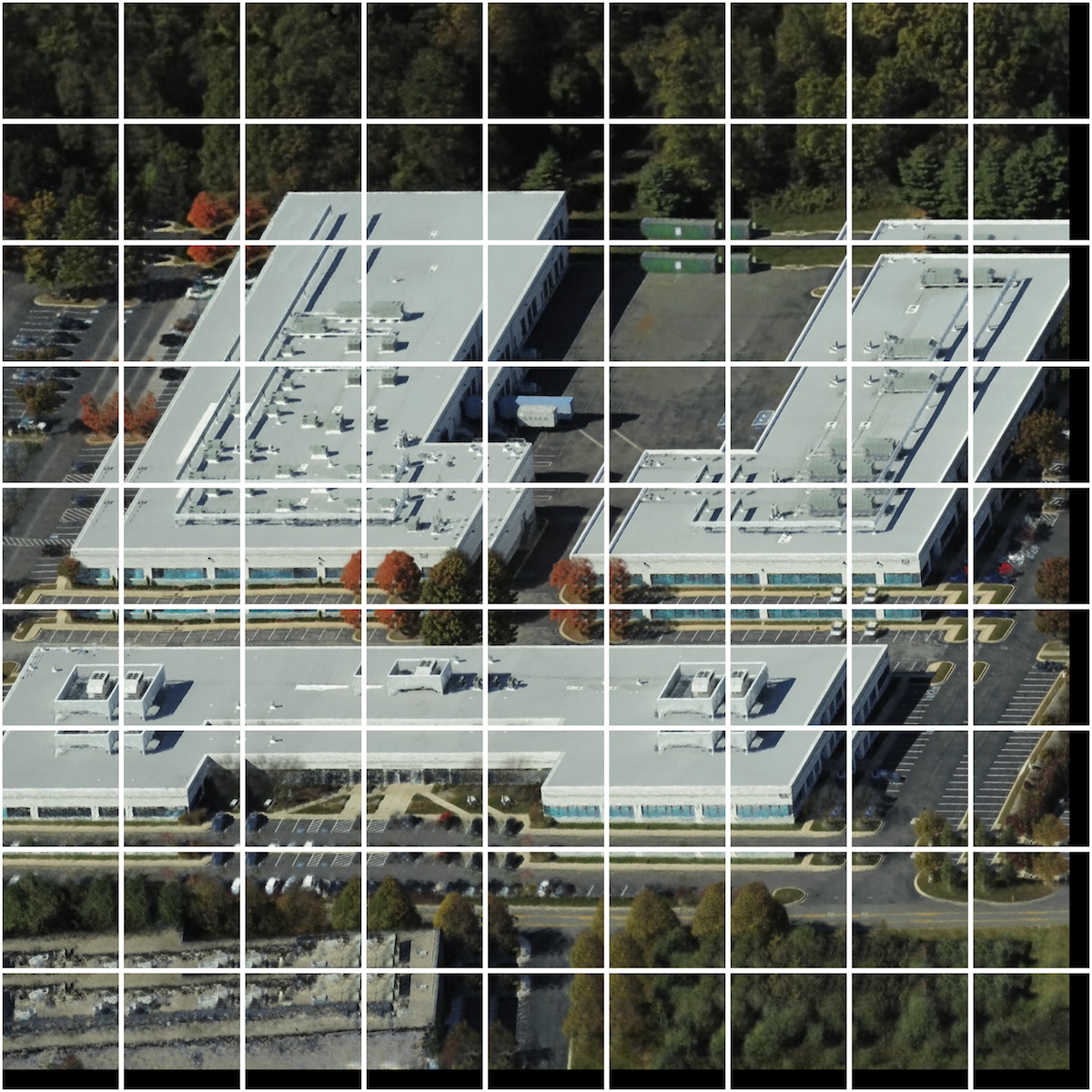}
    \caption{Overlapping tiles are extracted from the oblique airborne renderings for matching with ground imagery.}
    \label{fig:tiling_example}
\end{figure}

Given two orthophotos, one based on a (provided) georegistered satellite DSM and one rendered from the airborne NeRF-based model (\cref{sec:novelview} § Novel view rendering), we first compute the dense flow with RoMa between the two images (\cref{sec:robustcorr}).
Once we have thresholded away points that are poorly matched according to RoMa, as well as points that are poorly modeled by the NeRF-based aerial model (according to the co-rendered confidence map), we then compute the cyclic consistency of the dense matches (\cref{sec:denseflow}).
Matches that have poor cyclic consistency are very likely incorrect, and as such, are also discarded.
Incorrect matches can result in a poorly estimated 3D Similarity transform, which results in a poorly georegistered airborne model (with similar downstream effects for the ground-based georegistration).
Cyclically consistent matches, on the other hand, are more likely correct matches (\ie, they identify the same world point), and thus result in an accurate transform.

Dense matchers can result in a \emph{lot} of matches; however, many of these matches are redundant. For example, two matches that begin in neighboring pixels (and not on some 3D boundary) will likely be highly correlated.
To improve processing time, we reduce redundant information by setting a threshold of (up to) 5000 matches, which are randomly sampled from the remaining cyclically-consistent points.
This threshold was empirically chosen to balance the competing constraints of transform accuracy and processing time.
The 2D-to-2D matches are then lifted to 3D and a similarity transform is computed to bring the aerial model into alignment with the satellite DSM as described in \cref{sec:georegistration}.

\begin{figure}
    \centering
    \includegraphics[width=0.9\linewidth]{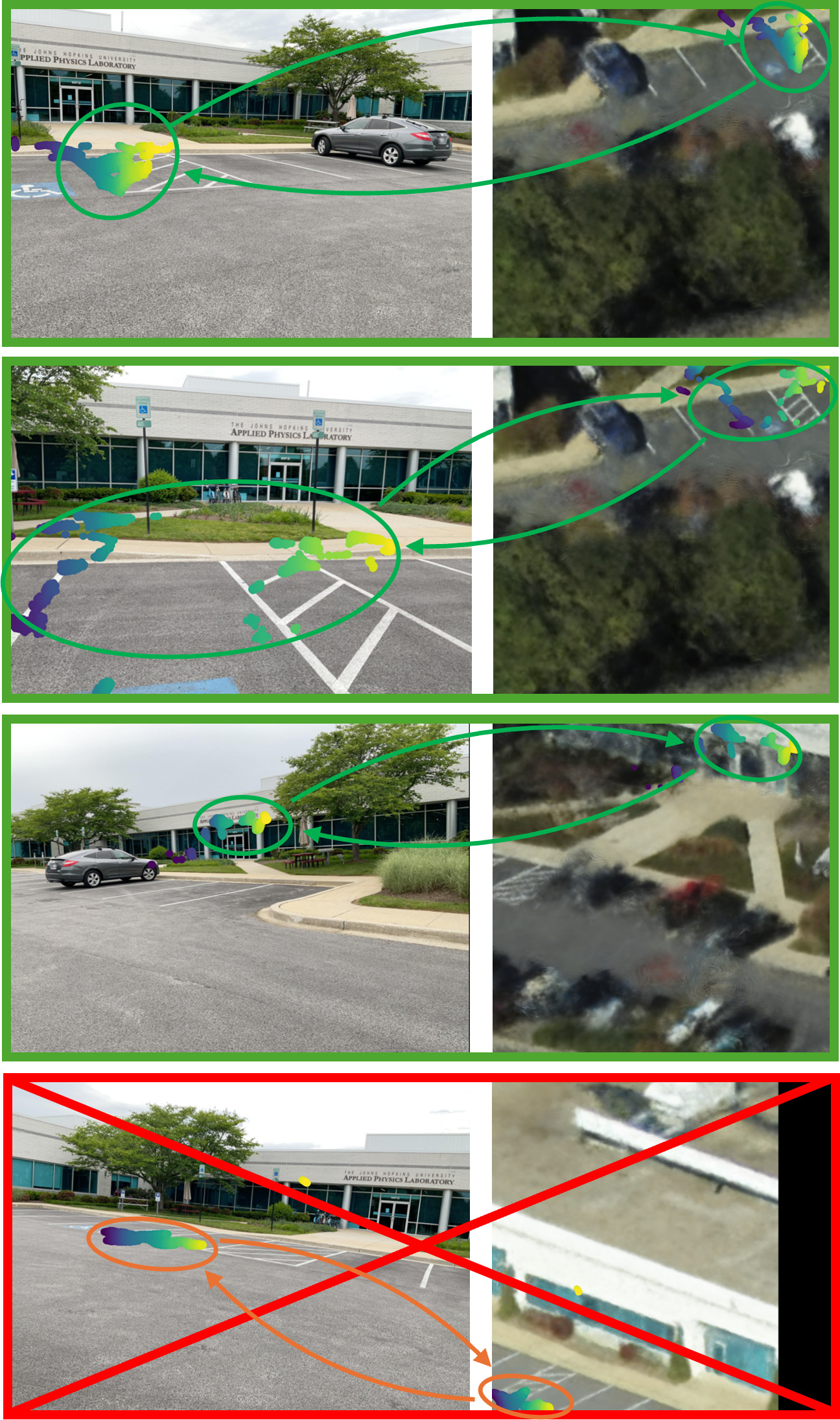}
    \caption{A consensus among the best matching tiles is taken, so a few incorrect ones do not influence the final transform.}
    \label{fig:ground_tile_matching}
\end{figure}

\subsection{Georegistration of ground imagery}
\label{sec:ground_georeg}

Metadata-free georegistration of ground-level imagery aims to transform a ground-based model into the georegistered airborne model's coordinate frame.
Similar to the process of registering the airborne imagery with satellite, the robust matching capabilities of RoMa are used to find cyclically consistent correspondences between NeRF-based renderings of the airborne site model and unmodified ground-level imagery \cref{sec:robustcorr}.
To mitigate the large difference in viewpoint between the airborne and ground images, the airborne model is rendered at oblique viewpoints to capture detail from the sides of buildings (as seen in \cref{fig:renders_grid}).
Since the ground imagery is assumed to have no associated metadata, the airborne model is rendered from eight different oblique viewpoints, each 45° apart in azimuth and all viewing the scene at a fixed 45° angle of depression.
This ensures good coverage of view angles present in the ground imagery.
The airborne renderings are generated at high resolution ($2048 \times 2048$ pixels) to preserve detail.
Cyclically consistent matching is then exhaustively performed between each ground image and the (tiled) airborne renders to identify the most likely matches.


\subsubsection*{Tiling of Airborne Renders}
The cyclically consistent matching pipeline will fail if the difference in viewpoint or field of view (FoV) between the images is too great.
Therefore, when attempting to match a single ground image to an airborne rendering viewing the entire scene, the airborne image is first subdivided into fixed-size tiles as shown in \cref{fig:tiling_example}.
In theory, the optimal size of these tiles could be computed per ground image if the metric scale of the ground model was known since the airborne site model is georegistered and has metric scale.
Initial experiments using metric depth estimation~\cite{yin2023metric} to estimate the scale of the ground models have shown promise but have not yet been integrated into the proposed method.
In the absence of scale information for the ground model, a fixed $300 \times 300$ pixel tile size is used with $25\%$ overlap between neighboring tiles.

\subsubsection*{Exhaustive Matching}
\label{sec:Exhaustive Matching}
Dense matching is performed between each ground image and all rendered airborne tiles (a total of $729$ tiles in our case).
For each ground image, 3D similarity transforms are computed for the top $k$ matching tiles, ranked by number of cyclically consistent matches.
The computed similarity transforms are evaluated based on their consistency with matched point correspondences from all other match pairs, and the transform with the most inliers is selected.
A final 3D similarity transform is computed using all inlier matches of the selected candidate transform.
Basing the computed transform on the largest consensus set of inliers ensures that the method is robust to (potentially many) incorrect matched pairs (see \cref{fig:ground_tile_matching}), which is critical for success.
We found thresholds of top $k=5$ tiles and inlier threshold of \qty{50}{\cm} worked well in practice and used these settings for all reported experiments.
Finally, Open3D's~\cite{Open3D} implementation of ICP~\cite{VanillaICP, ICPVariants} with the Tukey robust kernel \cite{ICPRobustKernels} is used to further refine the transform by aligning the full set of reconstructed ground and airborne points.

\begin{figure}
    \centering
    \includegraphics[width=\linewidth]{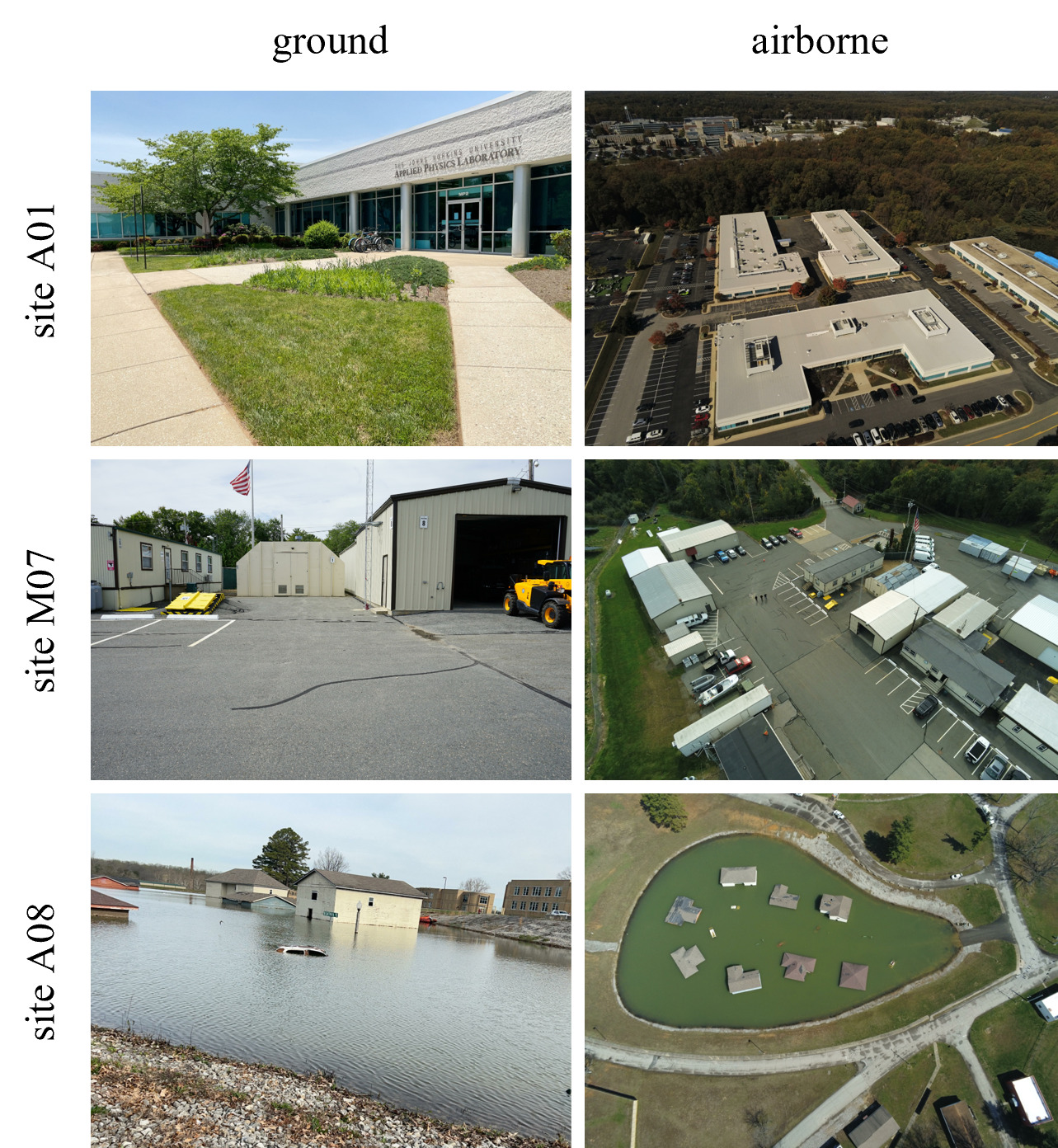}
    \caption{Sample ground and airborne imagery from sites A01, M07 and A08.}
    \label{fig:site_imagery_examples}
\end{figure}

\begin{table*}[t]
\centering
\begin{tblr}{
  width = \linewidth,
  colspec = {Q[90]Q[178]Q[127]Q[181]Q[162]Q[202]},
  cells = {c},
  vline{2-6} = {-}{},
  hline{2-4} = {-}{},
}
\textbf{Airborne} & \textbf{Mean Absolute Error (m)} & \textbf{M-dist w.r.t. DSM} & \textbf{Mean Relative Error (m)} & \textbf{Standoff Distance (m)} & \textbf{DSM uncertainty (CE90, LE90)}\\
Site A01 & 1.83 & 1.2 & 1.42 & 148 & (\qty{2.13}{\m}, \qty{3.82}{\m})\\
Site M07 & 0.83 & 0.4 & 0.67 & 113 & (\qty{3.11}{\m}, \qty{3.71}{\m})\\
Site A08 & 3.40 & 3.9 & 1.71 & 67 & (\qty{2.85}{\m}, \qty{3.84}{\m})
\end{tblr}
\caption{Results for airborne-to-satellite georegistration. ``M-dist'' indicates the Mahalanobis distance of the error relative to the DSM's reported covariance matrix. See \cref{sec:aerial_georeg_error} for discussion.}
\label{tab:airborne_results}
\end{table*}

\begin{table*}[t]
\begin{tblr}{
  width = \linewidth,
  colspec = {Q[85]Q[113]Q[150]Q[110]Q[149]Q[160]Q[171]},
  cells = {c},
  vline{2-7} = {-}{},
  hline{2-4} = {-}{},
}
\textbf{Ground} & \textbf{Number of Subgraphs} & \textbf{Mean Absolute Error (m)} & \textbf{M-dist w.r.t. DSM} & \textbf{Mean Relative Error (m)} & \textbf{Median Relative Error (m)} & \textbf{DSM uncertainty (CE90, LE90)}\\
Site A01 & 2 & 3.00 & 2.0 & 0.91 & 0.62 &  (\qty{2.13}{\m}, \qty{3.82}{\m})\\
Site M07 & 1 & 0.81 & 0.3 & 0.39 & 0.05 & (\qty{3.11}{\m}, \qty{3.71}{\m})\\
Site A08 & 2 & 6.03 & 2.9 & 4.50 & 1.2 & (\qty{2.85}{\m}, \qty{3.84}{\m})
\end{tblr}
\caption{Results for ground-to-airborne georegistration. ``M-dist'' indicates the Mahalanobis distance of the error relative to the DSM's reported covariance matrix. See \cref{sec:ground_georeg_results} for discussion.}
\label{tab:ground_results}
\end{table*}

\section{Results and analysis}
\label{sec:results}
We tested the proposed metadata-free georegistration method across three distinct test sites using datasets from the IARPA WRIVA program: A01 (office park), M07 (industrial setting) and A08 (flooded buildings within a pond), which are shown in \cref{fig:site_imagery_examples}.
The test data does not include ground control points for evaluation, so we instead evaluate the georegistration accuracy by comparing the estimated camera locations to their ground-truth locations.
The ground-truth camera locations are provided in geodetic coordinates, which we transform to a local Cartesian coordinate system and measure error as Euclidean ($L^2$) distance in meters.
We report the average georegistration error over all cameras for both the airborne and ground-level 3D models.

\begin{figure*}
    \centering
    \includegraphics[width=\linewidth]{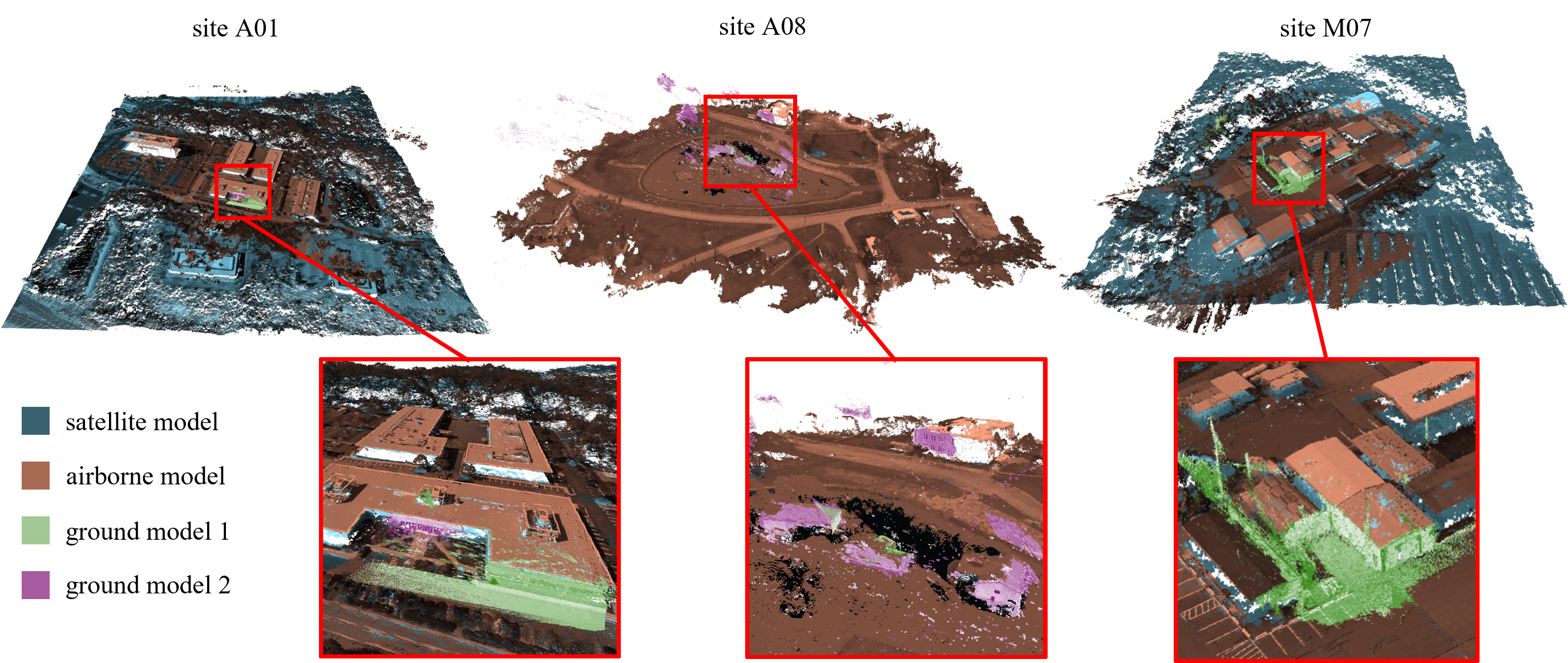}
    \caption{The satellite, airborne, and ground models from each site are georegistered together. The walls of the buildings are typically filled in by ground-level imagery, whereas the rooftops are from an airborne (or satellite) model. Satellite imagery (C) 2025 Maxar}
    \label{fig:three_sites}
\end{figure*}


\subsection{Airborne georegistration results}
\label{sec:aerial_georeg_error}


The airborne models align well visually and quantitatively.
\cref{fig:three_sites} shows the airborne model well aligned with the DSM (and the ground-level models). For example, there is no protrusion of the corners of the buildings due to poor alignment.
In addition to these qualitative examples, we also performed a quantitative evaluation, with results listed in \cref{tab:airborne_results}.
Site M07 has the lowest error of 0.83 meters and site A08 has the highest error of 3.40 meters (all 50 aerial cameras were reconstructed as part of the model for these two sites).
Note that the reported results include three sources of error: the geopositioning uncertainty of the reference DSM (which is a function of the source satellite imagery), the original COLMAP estimation of the (local) cameras and finally, the georegistration of the cameras.
Regarding the first source of error, the satellite DSM has an estimated uncertainty associated with it in the form of a $3\times3$ covariance matrix of a multivariate Gaussian distribution.
The Mahalanobis distance (``M-dist'')
of the average absolute error indicates how likely the observed error is based on the uncertainty of the DSM as described by its covariance matrix.
DSM uncertainty can account for most of the error present in sites M07 and A01; however, it is unlikely that the absolute error for site A08 is fully accounted for by the uncertainty from the DSM (Mahalanobis distance of 3.9). 
Note that DSM uncertainty is reported in \cref{tab:airborne_results} in the form of ``CE90'' and ``LE90'' indicating the \nth{90} percentile horizontal (circular) and vertical (linear) errors, respectively.


In order to further isolate the uncertainty in the geopositioning error, we also compute the \textit{relative} geopositioning error, which removes systematic bias by subtracting the mean error from estimated camera positions.
Shifting here is sufficient, as opposed to a rigid transform, because positioning errors in satellite imagery are primarily limited to translations~\cite{mundy_error_2022}.
As before, M07 has the lowest error, but now A08 and A01 have much more comparable error rates.

One plausible explanation for the smaller geopositioning error at site M07 is the shorter standoff distance (average distance from the sensor to the scene) as compared to sites A01 and A08.
If the geopositioning errors are normalized by the cameras' average distance to the center of the scene, we get very similar relative average error of $\sim 0.01$ meters per linear meter.

\subsection{Ground-level georegistration results}
\label{sec:ground_georeg_results}
Ground-level georegistration error is calculated identically to the airborne registration error (\cref{sec:aerial_georeg_error}) and is reported in \cref{tab:ground_results}.
Since the airborne site model is georegistered, the transform estimated in \cref{sec:Exhaustive Matching} brings the ground-based site model directly into a georegistered coordinate system which is used for error calculation.


Ground imagery from these sites often causes SfM pipelines such as COLMAP to produce several disjoint subgraphs when there is insufficient overlapping image content.
A 3D transform must therefore be computed and evaluated separately for each subgraph.
Unfortunately, the quality of these ground-based models is highly variable due to sparse imagery, numerous doppelgangers, and the inherent randomness of the COLMAP pipeline.
To filter out poor quality ground-based site models, an experimentally-determined minimum of nine cameras was required. 
Additionally, the total percentage of inlier correspondences (from \cref{sec:Exhaustive Matching}) was recorded from the final transform of each subgraph.
If a transform resulted in fewer than $20\%$ inliers, the model was considered unusable.
Using these two criteria, one subgraph was removed from the COLMAP-generated reconstructions for each of site A01 and A08.
Improved robustness in the initial SfM reconstruction and filtering algorithms remains a topic of future research.

After removing low-quality subgraphs, sites A01 and A08 each had two models (consisting entirely of ground imagery or security camera footage), while site M07 had only one.
The absolute and relative errors recorded in \cref{tab:ground_results} were averaged among all non-filtered subgraphs for each site.
The large variance in calculated mean errors is a result of errors in the underlying SfM models.
In particular, site A08 had significantly higher mean errors due to a single incorrect camera position in one of the COLMAP models.
This is evident by its significantly lower median relative error.
Meanwhile, the errors for site M07 were significantly lower overall due to a higher quality SfM model.
Sites A01 and A08 averaged around $11$ cameras per ground-based subgraph, whereas site M07 had $50$ cameras, which allowed COLMAP to more precisely predict the relative camera positions, and in turn resulted in much lower error.

\section{Conclusions and future work}
\label{sec:conclusion}
We proposed a metadata-free multi-domain georegistration method that leverages the photo-realistic novel view rendering capabilities of neural radiance fields and robust dense image matchers to mitigate the large viewpoint disparities between ground, airborne, and satellite imagery.
The proposed method demonstrates promising results in experiments on three visually distinct test sites.
The proposed method is robust but computationally intensive due to the dense matching performed across all pairs of rendered image tiles and input ground images, which have a relatively narrow field of view.
Future work includes mitigating this computational complexity by leveraging monocular metric depth estimation to predict scale a priori, and leveraging fast visual place recognition algorithms \cite{keetha2023anyloc} to reduce the number of candidate tiles for which matching is attempted.

\section*{Acknowledgment}
Supported by the Intelligence Advanced Research Projects Activity (IARPA) via Department of Interior/ Interior Business Center (DOI/IBC) contract number 140D0423C0035. The U.S. Government is authorized to reproduce and distribute reprints for Governmental purposes notwithstanding any copyright annotation thereon. Disclaimer: The views and conclusions contained herein are those of the authors and should not be interpreted as necessarily representing the official policies or endorsements, either expressed or implied, of IARPA, DOI/IBC, or the U.S. Government.

{\small
\bibliographystyle{ieee_fullname}
\bibliography{egbib}
}

\end{document}